\documentclass[10pt,twocolumn,letterpaper]{article}

\usepackage[pagenumbers]{iccv} 
\usepackage{makecell} 
\usepackage{multirow}
\definecolor{iccvblue}{rgb}{0.21,0.49,0.74}
\usepackage[pagebackref,breaklinks,colorlinks,allcolors=iccvblue]{hyperref}

\title{HumanDreamer-X: Photorealistic Single-image Human Avatars \\ Reconstruction via Gaussian Restoration}

\author{
    Boyuan Wang\textsuperscript{\rm 1, 2}\footnotemark[1]~,
    Runqi Ouyang\textsuperscript{\rm 1, 2}\footnotemark[1]~,
    Xiaofeng Wang\textsuperscript{\rm 1, 2}\footnotemark[1]~, 
    Zheng Zhu\textsuperscript{\rm 1}\footnotemark[1]~\footnotemark[2]~~,
    Guosheng Zhao\textsuperscript{\rm 1, 2} \\
    Chaojun Ni\textsuperscript{\rm 1, 3}, 
    Xiaopei Zhang\textsuperscript{\rm 4},
    Guan Huang\textsuperscript{\rm 1}, 
    Yijie Ren\textsuperscript{\rm 2},
    Lihong Liu\textsuperscript{\rm 2},
    Xingang Wang\textsuperscript{\rm 2}\footnotemark[2]\\
    \textsuperscript{\rm 1}GigaAI~~
    \textsuperscript{\rm 2}Institute of Automation, Chinese Academy of Sciences~~\\
    \textsuperscript{\rm 3}Peking University~~
    \textsuperscript{\rm 4}University of California Los Angeles
    \\
}

\begin{document}

\twocolumn[{
\vspace{-3em}
\maketitle
\vspace{-3em}
\begin{center}
\centering
\resizebox{0.9\linewidth}{!}{
\includegraphics{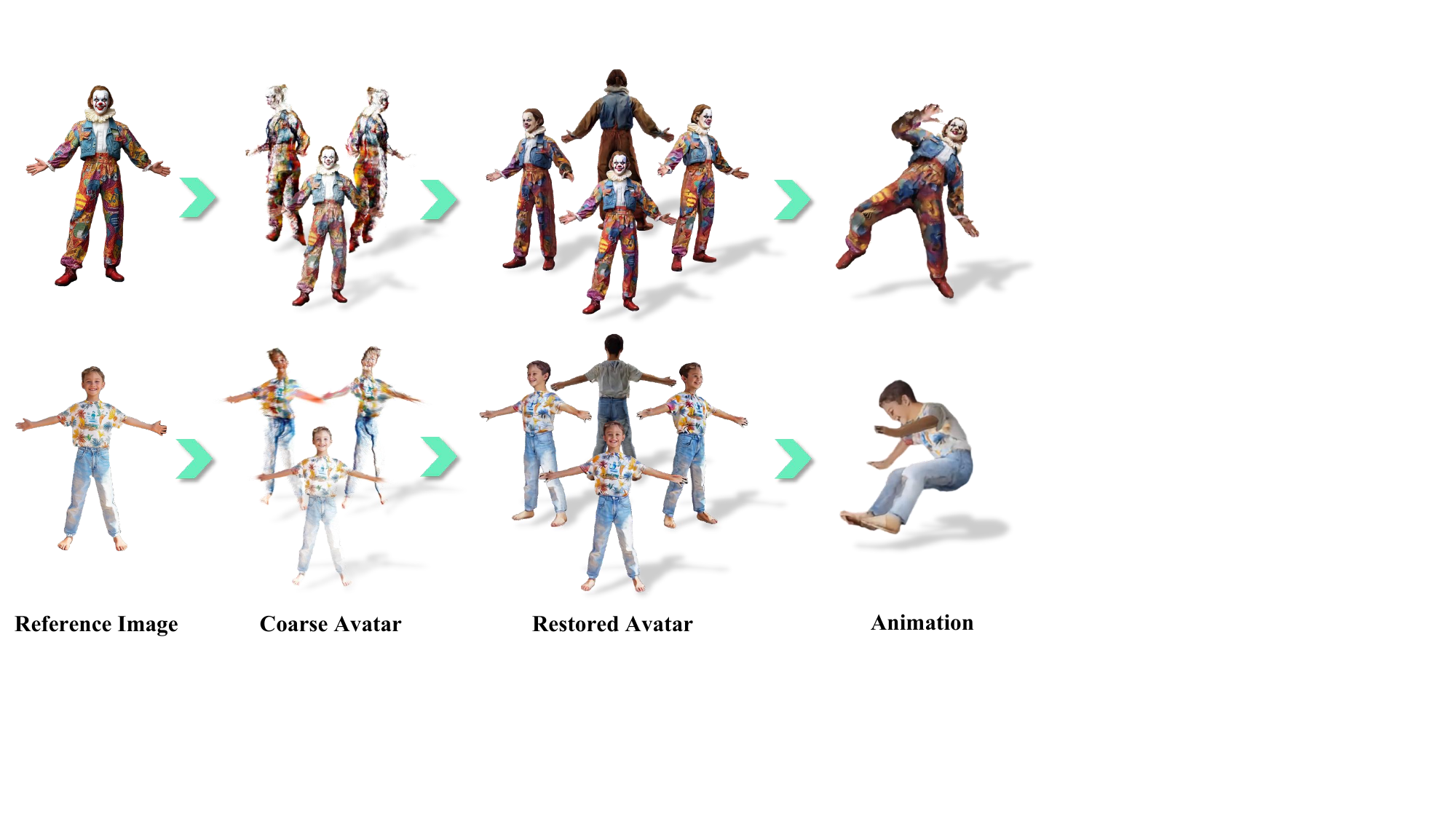
}}
\vspace{-0.5em}
\captionof{figure}{Illustration of \textit{HumanDreamer-X}. The pipeline initiates with a single-image reconstruction to generate a coarse 3D avatar, providing essential geometric and appearance priority for the restoration process. This approach facilitates the attainment of a higher-quality 3D avatar, suitable for subsequent animation tasks.}
\label{fig:main}
\end{center}}]
\renewcommand{\thefootnote}{\fnsymbol{footnote}}
\footnotetext[1]{
These authors contributed equally to this work. 
}
\footnotetext[2]{\mbox{Corresponding authors. zhengzhu@ieee.org, xingang.wang@ia.ac.cn.}}
\footnotetext[3]{Project Page: \url{https://humandreamer-x.github.io/}}

\begin{abstract}

Single-image human reconstruction is vital for digital human modeling applications but remains an extremely challenging task. Current approaches rely on generative models to synthesize multi-view images for subsequent 3D reconstruction and animation. However, directly generating multiple views from a single human image suffers from geometric inconsistencies, resulting in issues like fragmented or blurred limbs in the reconstructed models.
To tackle these limitations, we introduce \textbf{HumanDreamer-X}, a novel framework that integrates multi-view human generation and reconstruction into a unified pipeline, which significantly enhances the geometric consistency and visual fidelity of the reconstructed 3D models. In this framework, 3D Gaussian Splatting serves as an explicit 3D representation to provide initial geometry and appearance priority. Building upon this foundation, \textbf{HumanFixer} is trained to restore 3DGS renderings, which guarantee photorealistic results.
Furthermore, we delve into the inherent challenges associated with attention mechanisms in multi-view human generation, and propose an attention modulation strategy that effectively enhances geometric details identity consistency across multi-view.
Experimental results demonstrate that our approach markedly improves generation and reconstruction PSNR quality metrics by 16.45\% and 12.65\%, respectively, achieving a PSNR of up to 25.62 dB, while also showing generalization capabilities on in-the-wild data and applicability to various human reconstruction backbone models.
\end{abstract}

\section{Introduction}
Creating 3D human avatars is gaining increasing significance across various domains, including virtual reality, gaming, and film production. Among the numerous methods, creating from a single image represents a common, practical, and user-friendly approach. Nevertheless, constructing a versatile human avatar with diverse shapes, appearances, and clothing from just a single image remains a substantial challenge.

Traditional reconstruction methods based on mesh, Neural Radiance Fields (NeRF) \cite{nerf} and 3D Gaussian Splatting (3DGS) \cite{3dgs} enable multi-view reconstruction but are incapable of achieving single-image reconstruction \cite{SparseFusion, Text2Avatar, animatablegaussians, gaussianavatar, gauhuman, gart, instantavatar, humannerf}.

Despite these advancements, single-image reconstruction remains particularly challenging. To address these limitations, current approaches for single-view human reconstruction often integrate techniques from image or video generation \cite{sith, sifu, en3d, magicman, idol, charactergen, pshuman, anigs, gas}. A common strategy, as seen in methods like PSHuman \cite{pshuman} and AniGS \cite{anigs}, involves first generating multi-view images using generative models \cite{animateanyone, champ, mimicmotion, unianimate, animatex} and then performing subsequent reconstruction with mesh-based or Gaussian-based techniques. However, this decoupled paradigm heavily depends on the geometric and appearance consistency of the generative model. Any inconsistency in these aspects can lead to severe artifacts in the reconstruction stage, such as fragmented or distorted limbs.

To alleviate the aforementioned issues, we introduce \textit{HumanDreamer-X}, a novel framework that integrates multi-view human generation and reconstruction into a unified pipeline. This integration facilitates mutual enhancement between the two processes, offering supplementary information on geometry and appearance. Consequently, this significantly improves the geometric consistency and visual fidelity of the reconstructed 3D models, effectively alleviating problems related to fragmented and blurred limbs in the generated models.
As illustrated in Figure.~\ref{fig:framework}, our framework first utilizes 3DGS to reconstruct the human body from a single image and then renders multi-view images. Leveraging the geometric representation capability of 3DGS, these rendered multi-view images provide strong geometric and appearance priority for the subsequent generative process. Building upon this, \textit{HumanFixer} is introduced to refine the renderings, producing photorealistic images. Finally, these restored images are further utilized to guide 3DGS in reconstructing a high-quality human model.
Compared to decoupled approaches, our unified framework effectively bridges the gap between reconstruction and generation, producing higher-quality avatars suitable for diverse downstream applications.
Moreover, in the context of multi-view human generation, directly generating such videos often leads to blurriness because of temporal inconsistencies. To tackle this issue, we explore the inherent challenges associated with attention mechanisms within attention layers and propose a modulation strategy. This strategy effectively enhances geometric detail and identity consistency across multiple views, thus overcoming the limitations of traditional approaches.

Notably, our experiments demonstrate that our approach significantly improves the generation metrics by 16.45\% in PSNR and the reconstruction metrics by 12.65\% in PSNR. And also demonstrating its generalization capabilities on in-the-wild data and its adaptability to various human reconstruction backbone models.

The main contributions of this paper are summarized as follows:

\begin{itemize}
    \item We propose a novel framework, \textit{HumanDreamer-X}, for generating animatable avatars from a single-view image by coupling 3D reconstruction with video restoration. This unified approach significantly enhances the quality and consistency of reconstructed avatars compared to existing decoupled methods.
    
    \item We identify and address attention-related deficiencies in the generation of multi-view videos, which often lead to inconsistencies and blurriness. To mitigate these issues, we introduce an attention correction module that refines the temporal attention mechanism, thereby improving the quality of restored videos and ensuring better geometric and identity preservation.

    \item Extensive experimental results demonstrate the superior performance of \textit{HumanDreamer-X} across multiple datasets. Specifically, our method achieves higher fidelity in avatar reconstruction, showcasing its practical applicability and efficiency. It also demonstrates generalization capabilities on in-the-wild data and is applicable to various human reconstruction backbone models.
\end{itemize}

\section{Related Work}

\subsection{Single-image Human Recosntruction}

With the advancement of 3D representation methods such as mesh \cite{Hu_2021_CVPR, meshanything}, NeRF \cite{nerf,mipnerf,zipnerf,mipnerf360,10229247}, and 3DGS \cite{3dgs,mipgs,lu2024scaffold,deformablegs}, human reconstruction has seen significant improvements \cite{humanrf,hugs, 9787789, evahuman,avatarrex,3dgsavatar,eva3d,humansplat,ag3d}. However, these methods typically require a large number of multi-view images or videos for reconstruction, which limits their applicability. In contrast, Single-image human reconstruction is a more flexible task, aiming to reconstruct a 3D human model from just one image \cite{anigs,pshuman,charactergen,idol,sith,sifu,magicman,en3d}. However, it is inherently an ill-posed problem, presenting considerable challenges. 

Current single-image human reconstruction methods integrate techniques from generative approaches \cite{sd,svd,controlnet,yang2024cogvideox,zhu2024sora}. These methods first generate unseen viewpoints \cite{sith} or multi-view images \cite{anigs,pshuman,en3d,magicman,charactergen}, followed by employing 3D representations to reconstruct the human. To ensure geometric consistency, methods such as SMPL \cite{anigs,pshuman}, masks \cite{sith}, segmentation maps \cite{magicman}, and pose estimation \cite{charactergen} are utilized to drive the generation process. However, these generate-then-reconstruct approaches heavily rely on the geometric and appearance consistency of the generated images, and inconsistencies can lead to issues like fragmented limbs and blurriness in the reconstructed models.
Recently, there have been attempts \cite{idol} to infer 3DGS parameters directly from a single image using feed-forward networks, but this requires extensive datasets for training. The approach most closely related to ours is SIFU \cite{sifu}, which refines the coarse texture of reconstructed models using generative models. However, SIFU uses a frozen image generation model for frame-by-frame refinement, making it difficult to maintain inter-frame consistency.

\begin{figure*}[htbp]
    \centering
    \resizebox{0.9\textwidth}{!}{%
        \includegraphics{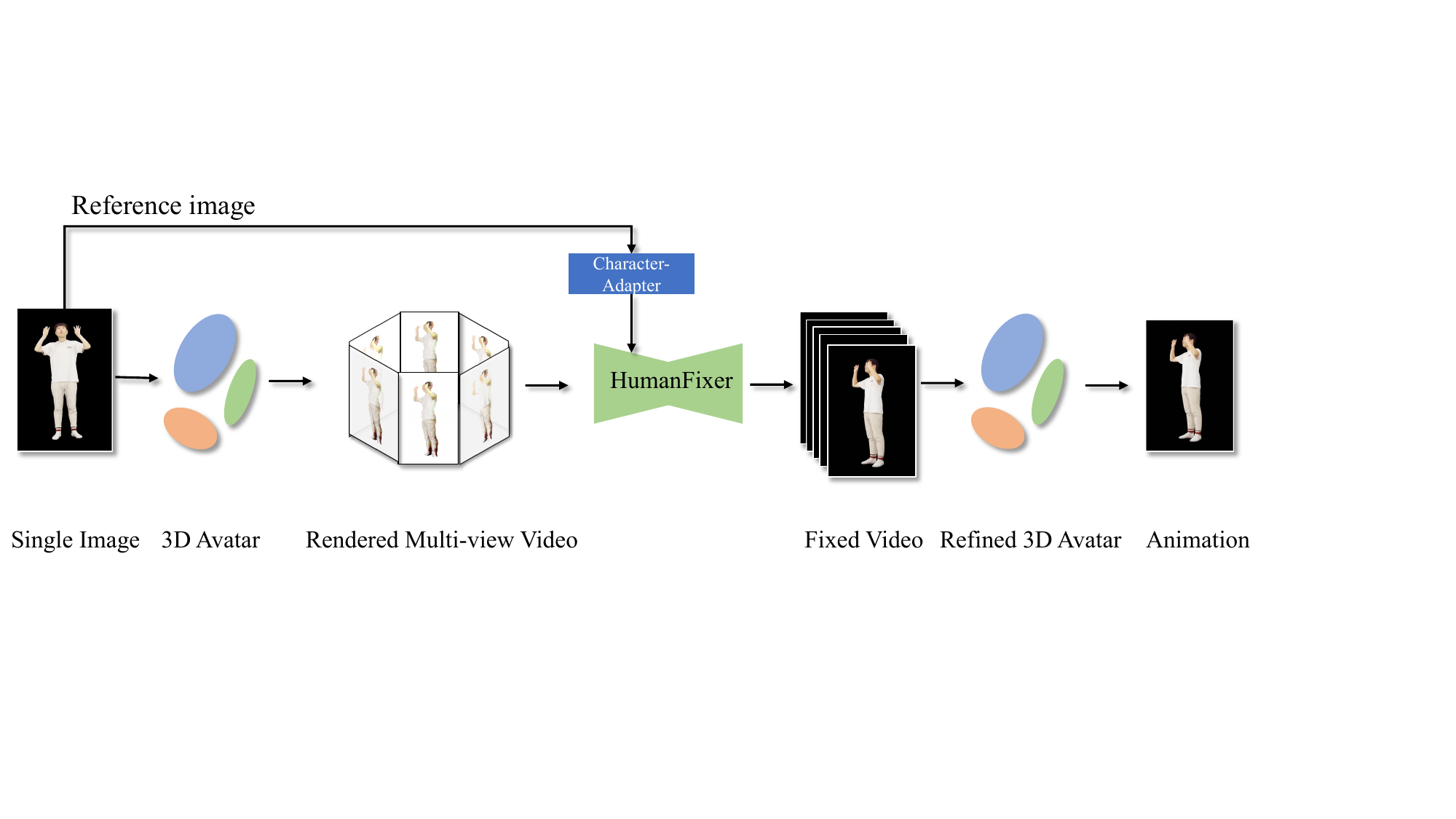}
    }
    \caption{Overall framework of the proposed \textit{HumanDreamer-X}. The process begins by initializing a coarse 3DGS avatar using a reference image. A rendered video serves as a , providing geometric and appearance priors. Subsequently, \textit{HumanFixer} performs video restoration, wherein an attention modulation is employed to enhance video consistency. Throughout this process, the restored video is used to continuously update the 3DGS model, ultimately resulting in a refined 3DGS avatar.}
    \label{fig:framework}
\end{figure*}

\subsection{Controllable Human Generation}

With the rapid advancements in image and video generation \cite{svd,controlnet,yang2024cogvideox,ma2024latte,eg3d,worlddreamer,yan2021videogpt,videoldm}, diffusion-based human generation models \cite{ipadapter, animateanyone, champ,mimo, animateanyone2, mimicmotion,animatex,emo,followyouremoji,aniportrait,10543061} have gained increasing attention. Among these, Animate Anyone \cite{animateanyone} stands out as a representative work, which constructs a 3D U-Net upon Stable Diffusion \cite{sd}  for modeling temporal information, and incorporates skeleton information as driving signals for controlling motion generation.

To further enhance temporal consistency, MimicMotion \cite{mimicmotion}  leverages pre-trained video generation model SVD \cite{svd} to improve the coherence of generated sequences. In contrast, UniAnimate \cite{unianimate} employs MAMBA-based \cite{mamba} techniques for enhanced temporal modeling.
Additionally, several approaches explore different motion control signals. For instance, DisCo \cite{disco} and MagicAnimate \cite{magicanimate} utilize DensePose \cite{densepose} as a representation of the human body \cite{disco,magicanimate}, while Champ \cite{champ} integrates multi-modal information including depth, normal, and semantic signals derived from the 3D parametric human model SMPL \cite{smpl}.
Despite their ability to generate photorealistic frames, these models often struggle to fully ensure geometric and appearance consistency across frames. This inconsistency poses significant challenges for subsequent reconstruction tasks.

\section{Method}

\subsection{Preliminary}
\noindent\textbf{3D Gaussian Splatting.}
3DGS represents a voxel-based rendering technique for scene representation, where the core concept involves modeling the scene using an optimized set of 3D Gaussian distributions $\mathcal{G} = \{ \mathcal{N}(\mathbf{x}_i, \Sigma_i) \}_{i=0}^{N}$. Each Gaussian distribution $\mathcal{N}$ is parameterized by its position $\mathbf{x}_i \in \mathbb{R}^3$, covariance matrix $\Sigma_i$ (which defines the ellipsoidal shape), and radiance attributes such as color $\mathbf{c}_i$ and opacity $\alpha_i$. During rendering, these Gaussian distributions are projected onto the image plane. The contribution of each pixel $\mathbf{u}$ is computed via radial basis functions defined as $w_i(\mathbf{u}) = \alpha_i \exp\left(-\frac{1}{2} (\mathbf{u} - \mathbf{u}_i)^\top \Sigma_i^{-1} (\mathbf{u} - \mathbf{u}_i)\right)$, where the weight $w_i$ decays with distance from the center of the Gaussian. This process is achieved through differentiable rasterization, which integrates depth sorting and weighted blending ($\sum_i w_i \mathbf{c}_i / \sum_i w_i$), thereby facilitating gradient-based optimization for high-quality scene reconstruction.

\noindent\textbf{Video Diffusion Model.}
Video Diffusion Models (VDM) are generative frameworks that extend diffusion-based image synthesis to dynamic scenes by modeling sequential data in a latent space. The core principle involves a forward diffusion process that gradually corrupts video data $\mathbf{x}_0$ into noise via $T$ steps: $q(\mathbf{x}_t | \mathbf{x}_{t-1}) = \mathcal{N}(\mathbf{x}_t; \sqrt{1-\beta_t}\mathbf{x}_{t-1}, \beta_t \mathbf{I})$, where $\beta_t$ controls noise injection.
VDM enhances temporal coherence by integrating spatiotemporal layers into a pre-trained latent space, enabling high-fidelity video generation from images or text while leveraging efficient VAE-based compression. As a reverse denoising network it learns to approximate the posterior $p_\theta(\mathbf{x}_{t-1} | \mathbf{x}_t)$, typically parameterized as $\mathbf{x}_{t-1} = \mu_\theta(\mathbf{x}_t, t) + \sigma_t \epsilon$ ($\epsilon \sim \mathcal{N}(0, \mathbf{I})$), reconstructing coherent video frames through iterative refinement.

\subsection{Overall Framework of HumanDreamer-X}

Traditional human reconstruction methodologies \cite{animatablegaussians, gaussianavatar, gauhuman, gart, instantavatar, humannerf} encounter substantial challenges due to their reliance on multi-view images. Recent advances \cite{sith, sifu, en3d, magicman, idol, charactergen, pshuman, anigs, gas} have endeavored to address this limitation by leveraging generative priors to compensate for the absence of invisible views. Nevertheless, the decoupling of generation from reconstruction has resulted in a deficiency of geometric consistency among the generated multi-view images. 
Our proposed framework, \textit{HumanDreamer-X}, integrates reconstruction and generation into a unified pipeline. In this pipeline, the reconstruction step provides initial geometry and appearance priorities, and the generative model then refines the reconstructions by restoring coarse renderings. The overall framework is illustrated in Fig.~\ref{fig:framework}. 

Specifically, we first use the 3DGS model \cite{animatablegaussians, gaussianavatar} to reconstruct an avatar $\mathcal{A}_c$ from a single reference image $\mathcal{I}_R$:
\begin{equation}
    \mathcal{A}_c = \text{3DGS}(\mathcal{I}_R, \theta_\text{SMPL}),
\end{equation}
Following previous avatar reconstruction models, the initial point cloud for 3DGS is derived from the estimated Skinned Multi-Person Linear (SMPL) $\theta_\text{SMPL}$. This ensures that even a single image can reconstruct coarse human geometry and appearance, providing a basic priority for subsequent refinement.
Then, a multi-view video $\mathcal{V}_c$ of the avatar is rendered using the trained 3DGS avatar:
\begin{equation}    
    \label{eq:render}
    \mathcal{V}_c = \{\mathcal{A}_c(d_i) \mid i = 1, 2, \ldots, n\},
\end{equation}
where $d_i$ represents a specific horizontal angle, and $\mathcal{A}_c(d_i)$ renders the avatar at that angle.
Due to the limited availability of only one viewpoint, the resulting model $\mathcal{A}_c$ provides only basic geometric and appearance priority of the avatar, leaving issues such as blurriness and artifacts in unseen views unresolved. To address these issues, we introduce \textit{HumanFixer}, a video generation model designed to restore details in the initial 3DGS renderings.

\textit{HumanFixer} is built upon the pretrained video diffusion model \cite{svd}, utilizing the coarse video $\mathcal{V}_c$ and the reference image $\mathcal{I}_R$ as conditions to generate multi-view refined videos (see Sec.~\ref{sec:humanfixer} for more details). The refined videos $\mathcal{V}_r$ capture the textures and geometry present in the reference image $\mathcal{I}_R$, making them suitable for modeling a refined human avatar $\mathcal{A}_r$.

This approach leverages the geometry and appearance priority provided by 3DGS and exploits the temporal consistency inherent in video generation models, ensuring enhanced multi-view consistency in the repaired avatar. Additionally, to address blurriness in multi-view video generation, we investigate attention mechanisms and propose an attention modulation strategy (see Sec.~\ref{sec:attn_mask} for more details). The refined video, enhanced through this strategy, is then utilized to optimize the human avatar.

\subsection{Training and Inference of HumanFixer}
\label{sec:humanfixer}
Reconstructed avatars derived directly from single-image inputs exhibit issues such as blurring at invisible views. To address these problems, we introduce \textit{HumanFixer} for refining coarse avatars. This section will present the methodologies for the training and inference of \textit{HumanFixer}.

\begin{figure}[htbp]
    \centering
    \resizebox{\columnwidth}{!}{
        \includegraphics{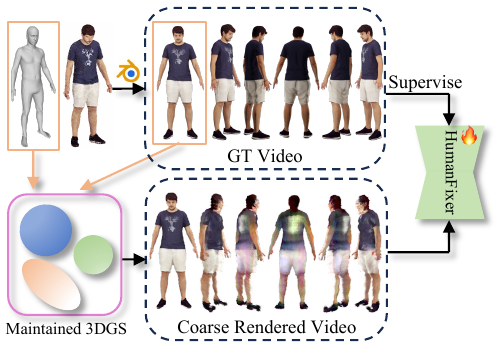}
    }
    \caption{The creation of the dataset for training \textit{HumanFixer}. First, we use Blender to render scans and obtain the ground truth video. Next, we employ the frontal image and its corresponding SMPL prior to reconstruct a coarse 3DGS model, followed by rendering multi-view videos. This process yields paired video data for training.}
    \label{fig:dataset}
\end{figure}

\noindent\textbf{Training.}
\textit{HumanFixer} hinges on constructing a coarse-refined pair dataset. In this section, we propose a concrete pipeline for dataset collection, as illustrated in Fig.~\ref{fig:dataset}. The steps are as follows: Initially, we utilize Blender to render a multi-view video $\mathcal{V}_{\text{gt}}$ from each 3D scan, serving as the ground truth video.  Subsequently, we leverage the GS model to reconstruct human avatar from single image, and use Eq.~\ref{eq:render} to render multi-view coarse images $\mathcal{V}_c$, paired with their corresponding ground truth videos $\mathcal{V}_{\text{gt}}$, form the repair dataset. This dataset is designed to refine avatar reconstructions by providing examples of both low-quality and high-quality renderings, allowing the \textit{HumanFixer} model to learn how to enhance the coarse images into refined, detailed videos.

The architecture of \textit{HumanFixer} is illustrated in Fig.~\ref{fig:framework}. It employs SVD \cite{svd} as the backbone and leverages the low-quality video to guide the generation process. 
Additionally, it integrates a CLIP \cite{clip} encoder to inject reference image information through cross-attention mechanisms.

During the training of \textit{HumanFixer}, we first feed the coarse video $\mathcal{V}_c$ into a Variational Autoencoder (VAE) \cite{vae} encoder $\mathcal{E}$ to obtain the latent feature $z_c = \mathcal{E}(\mathcal{V}_c)$. Then, $z_c$ is used as a condition, providing both geometric and appearance priority. It is concatenated with $z_{gt} = \mathcal{E}(\mathcal{V}_{gt})$, serving as the input to the model. To maintain identity consistency across multiple views, we utilize a reference image $\mathcal{I}_R$ as a source of identity information. The face embedding of the reference image $\mathcal{M}_f \in R^{n_t \times C}$ is extracted using an existing facial embedding extraction model from $\mathcal{I}_R$. $n_t$ denotes number of tokens and $C$ means dimension of cross-attention. The model's output is:
\begin{equation}
\epsilon_{\text{target}} = h_{\theta}(z_{gt}^t, z_c, \mathcal{M}_f),
\label{eq:formulaB}
\end{equation}
Where $h_\theta$ denotes the \textit{HumanFixer} module with parameters $\theta$, $z_{gt}^t$ represents the latents of the ground truth video $\mathcal{V}_{gt}$ at the noising time step $t$, and $z_c$ signifies the latents of the coarse video $\mathcal{V}_c$. 
As described in \cite{svd}, we use a predicted target loss \cite{svd} for optimizing the \textit{HumanFixer} model. 

\noindent\textbf{Inference.}
After training \textit{HumanFixer}, the coarse video $\mathcal{V}_c$ to be refined and the reference image $\mathcal{I}_R$ are used as inputs to obtain the refined video $\mathcal{V}_r$. 
\begin{equation}
\mathcal{V}_r = \text{EDM}(h_{\theta}(z^T, z_c, \mathcal{M}_f)), 
\label{eq:formulaC}
\end{equation}
where $z^T$ is the initial latent noise, and we use EDM scheduler \cite{svd} for denoising.

\begin{figure}[htbp]
    \centering
    \resizebox{\columnwidth}{!}{
        \includegraphics{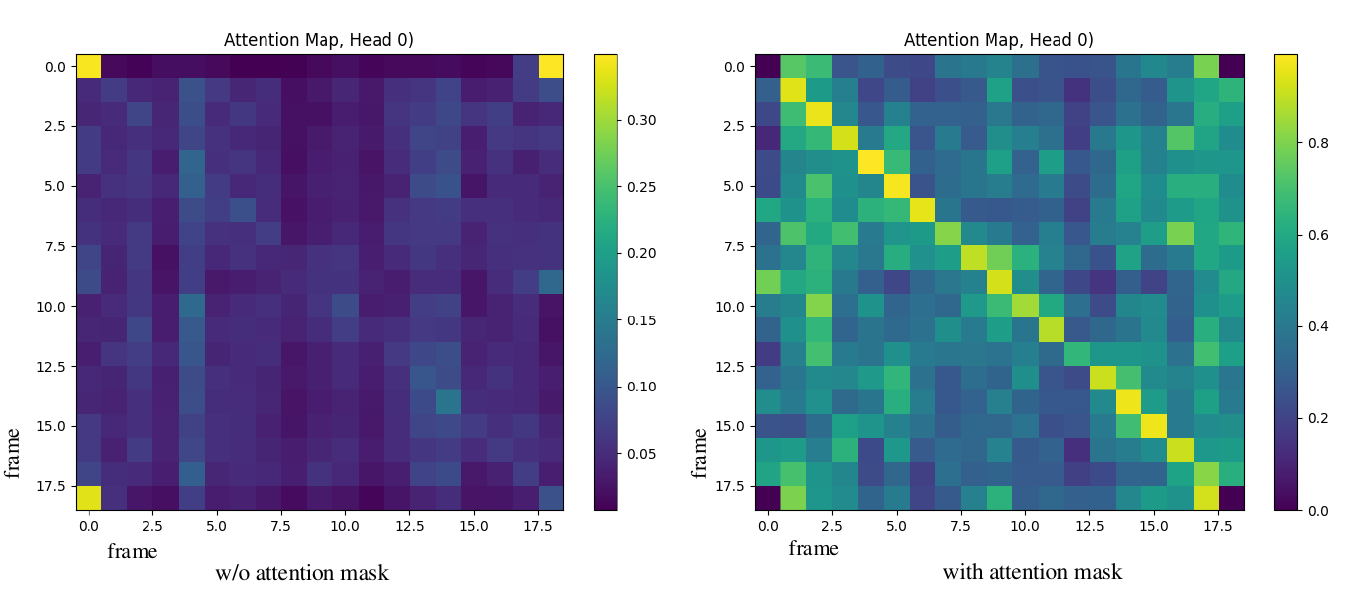}
    }
    \caption{Attention weights visualization. The left and right sides show the head 0 attention weights at the temporal self-attention stage for training on cyclic videos without and with an attention modulation, respectively. Brighter colors indicate higher weights.}
    \label{fig:attn_mask}
    \vspace{-0.3cm}
\end{figure}

\begin{table*}[htbp]
    \centering
    \caption{Multi-view video generation comparison of other SOTA methods. \textbf{Bold} indicate the best result.}
    \begin{tabular}{lccccccc}
        \toprule
        Method & Testset & PSNR $\uparrow$ & SSIM $\uparrow$ & LPIPS $\downarrow$ & FID $\downarrow$\\
        \midrule
        PSHuman \cite{pshuman} & \multirow{3}{*}{CustomHumans} & 21.998 & 0.826 & 0.1945 & 103.808 \\
        Champ \cite{champ} &   & 20.228 & \textbf{0.889} & 0.2438 & 115.031 \\
        HumanFixer (Ours) &     & \textbf{25.618} & 0.882 & \textbf{0.0687} & \textbf{87.149}\\
        \midrule
        Champ \cite{champ} & \multirow{2}{*}{THuman2.1} & 17.547 & 0.859 & 0.2701 & 129.629\\
        HumanFixer (Ours) &  & \textbf{23.741} & \textbf{0.889} & \textbf{0.0720} & \textbf{94.570} \\
        \bottomrule
    \end{tabular}
    \label{tab:gen_comp}
\end{table*}

\subsection{Attention Mechanism Analysis}
\label{sec:attn_mask}
Generating multi-view videos differs from generating standard videos due to the distinct temporal relationships involved. In typical videos, adjacent frames are strongly correlated, with correlation decreasing as the distance between frames increases. However, in multi-view videos, which encompass a full circle of views, the final frames have a strong relationship with the initial frames. Given that our model is fine-tuned on SVD, which is pretrained on standard videos, the pretrained parameters align more closely with the assumption of strong correlations between adjacent frames.

For a multi-view video with a total of \(N\) views, we define a non-cyclic video as one where each frame corresponds sequentially to views \(0, 1, ..., N-1\). Conversely, a cyclic video involves frames corresponding to views \(0, 1, ..., N-1, 0\), effectively looping back to the initial view.

Compared to non-cyclic videos, cyclic videos append the first frame to the end, making the training data more suitable for models pretrained on the assumption of strong correlations between adjacent frames. This should theoretically enhance overall consistency. However, during training, we observed that directly training on non-cyclic videos allows view \(0\) to develop stronger associations with views \(N-1\), \(N-2\), and \(N-3\). Nevertheless, a significant discontinuity was noted between view \(N-1\) and view \(0\), whereas the difference between view \(1\) and view \(0\) was relatively minor.

To address this issue, we further analyzed the temporal self-attention mechanism within the model. As illustrated in the left panel of Fig.~\ref{fig:attn_mask}, the attention weights for the starting and ending frames are significantly higher than those for intermediate frames. We hypothesize that this occurs because, in cyclic videos, the first and last frames are identical, leading to naturally higher attention weights compared to those calculated between different frames. This phenomenon suppresses information flow among intermediate frames.

In summary, while cyclic videos aim to improve consistency by leveraging adjacency assumptions, they introduce challenges related to discontinuities and uneven attention weight distribution, necessitating further refinement of the temporal self-attention mechanism.
To mitigate this issue, we apply an attention modulation, as shown in Fig.~\ref{fig:framework}, which ensures that the first and last frames do not receive attention from the model, either from themselves or from each other. 

Formally, let the attention mask $ M \in \mathbb{R}^{(N+1) \times (N+1)} $ be defined as:

\vspace{-0.5cm}
\begin{equation}
\begin{minipage}{\dimexpr\columnwidth-2\parindent\relax}
\centering
\resizebox{\linewidth}{!}{
$
M(i,j) = 
\begin{cases} 
-\infty, & \text{if } (i,j)=(0,0),(0,N),(N,0),(N,N) \\
0, & \text{otherwise}.
\end{cases}
$
}
\end{minipage}
\end{equation}
Then, this attention mask is added to every temporal self-attention module in the model:
\begin{equation}
\text{Attn}(Q, K, V, \mathcal{M}) = \text{softmax}\left(\frac{QK^T}{\sqrt{d_k}} + \mathcal{M}\right)V,
\label{eq:attention}
\end{equation}
where $Q$ for query, $K$ for key, $V$ for value, $d_k$ the dimension for scaling down the dot product results, $\mathcal{M}$ denotes the attention mask on temporal self attention.

This formulation effectively suppresses attention weights for the first and last frames, preventing the model from being unduly influenced by these boundary frames during video generation. Fig.~\ref{fig:attn_mask} illustrates that, after training on cyclic videos with the attention modulation, the attention weights shift from being predominantly focused between the 0th and $N$th frames to resulting in a more evenly distributed attention pattern.

\section{Experiments}
This section outlines our experimental framework, including the datasets, implementation specifics, and evaluation criteria. We then provide both quantitative and qualitative results to highlight the outstanding performance of the proposed \textit{HumanDreamer-X}.

\begin{figure*}[htbp]
    \centering
    \resizebox{\textwidth}{!}{
        \includegraphics{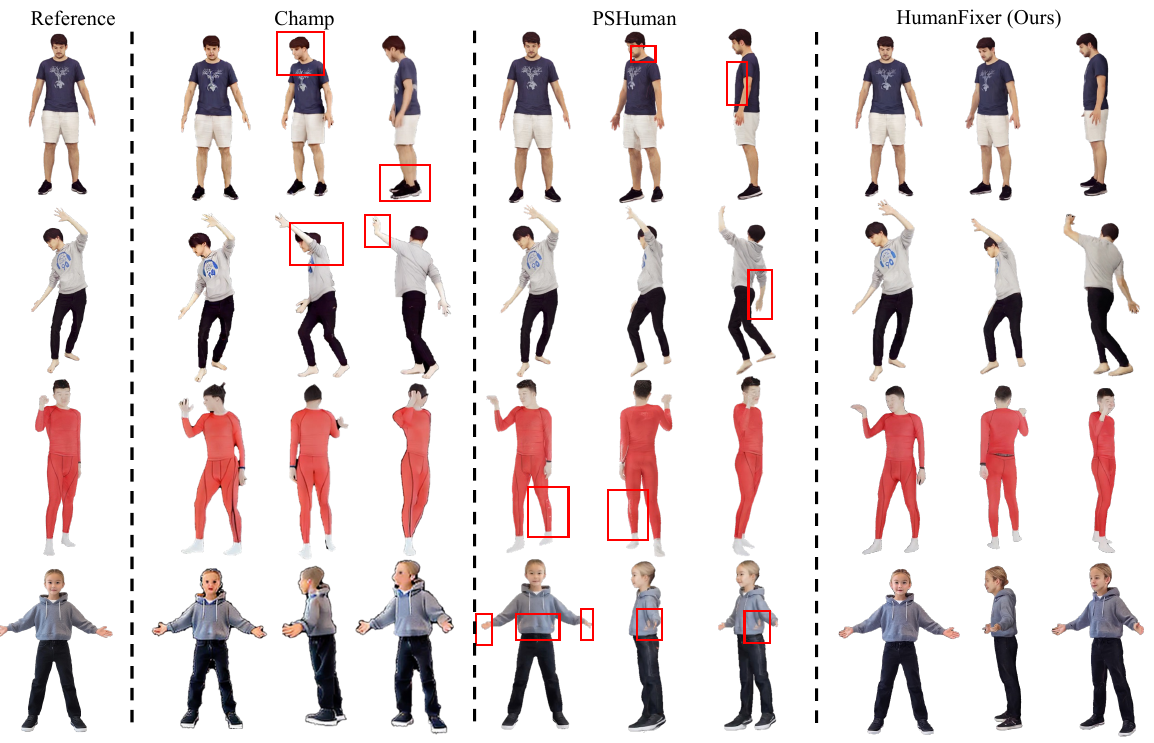}
    }
    \caption{Comparison of generation with SOTA methods. Note that PSHuman's training dataset contains all of the CustomHumans. Best viewed with zoom-in.}
    \label{fig:gen_comp}
\end{figure*}

\subsection{Experiment Setup}
\noindent\textbf{Dataset.}
Our experiments are conducted on a variety of datasets. To ensure that the human subject occupies a significant portion of the frame during training, we filter out instances where the arms are excessively spread, which would otherwise result in an overabundance of background content. The final training set comprises 388 scans from CustomHumans \cite{customhumans} and 1929 scans from THuman2.1 \cite{function4d}. For testing, we use 39 samples from \cite{customhumans}, 32 samples from \cite{function4d}. All training data is standardized to a resolution of $960 \times 640$, with cyclic video sequences consisting of 19 frames (18 multi-view frames plus the first frame repeated).

To ensure the reconstruction of a complete avatar, it is stipulated that the input to \textit{HumanFixer} must include facial information as identity cues. For the THuman2.1 dataset \cite{function4d}, facial detection is performed using InsightFace \cite{arcface}, and the image with the largest facial area is designated as the reference image. For the CustomHumans dataset \cite{customhumans}, the first frame is directly selected as the reference image.

\noindent\textbf{Baselines.}
For multi-view video generation, we utilize PSHuman \cite{pshuman} and Champ \cite{champ} as baselines. 
For 3D avatar reconstruction, we compare our method with a variety of existing single-image human reconstruction approaches. These include mesh-based methods such as PIFU \cite{pifu}, PaMIR \cite{pamir}, SiTH \cite{sith}, and PSHuman \cite{pshuman}, as well as recent feed-forward 3D Gaussian Splatting methods like IDOL \cite{idol} and LHM \cite{lhm}. In contrast to these end-to-end approaches, our framework adopts a two-stage generation-then-reconstruction pipeline. To validate the effectiveness of our framework across different generation strategies, we use either Champ \cite{champ} or \textit{HumanDreamer-X} for the initial human prior generation stage, while keeping the subsequent 3D reconstruction stage fixed—implemented on two distinct 3DGS backbones: Animatable Gaussians~\cite{animatablegaussians} and GaussianAvatar~\cite{gaussianavatar}.

\subsection{Main Results}
\noindent\textbf{Quantity comparison results on multi-view generation.}
We compare the multi-view generation performance of \textit{HumanFixer} with several SOTA methods, and the results are presented in Tab.~\ref{tab:gen_comp}. The findings indicate that \textit{HumanFixer} achieves superior video consistency. Fig.~\ref{fig:gen_comp} visually illustrates the enhanced multi-view consistency produced by our method.

\begin{table*}[!t]
    \centering
    \caption{3D reconstruction comparison. * denotes the metric is from PSHuman\cite{pshuman}.}
    \begin{tabular}{lcccccc}
        \toprule
        Method & Testset & PSNR $\uparrow$ & SSIM $\uparrow$ & LPIPS $\downarrow$ & FID $\downarrow$\\
        \midrule
        PSHuman  &\multirow{7}{*}{CustomHumans}& 20.089 & 0.8439 & 0.1770 & 87.816\\
        LHM && 20.153 & 0.9122 & 0.1121 & 110.341 \\
        IDOL && 18.333 & 0.9209 & 0.0882 & 187.710\\
        Champ with GaussianAvatar && 19.673 & 0.8789 & 0.2643 & 164.554 \\
        \textbf{\textit{HumanDreamer-X}} with GaussianAvatar && \textbf{23.639} & 0.9100 & 0.2427 & 114.804 \\
        Champ with Animatable gaussians &&16.853 & 0.9157 & 0.1251 & 122.752 \\
        \textbf{\textit{HumanDreamer-X}} with Animatable gaussians && 22.631 & \textbf{0.9458} & \textbf{0.0729} & \textbf{71.250}\\
        \midrule
        PIFU$^*$ &\multirow{10}{*}{THuman2.1}& 19.3957 & 0.8327 & 0.1001 & - \\
        PaMIR$^*$ && 19.4130 & 0.8324 & 0.0988 & - \\
        SiTH$^*$ && 18.458 & 0.8200 & 0.1004 & - \\
        PSHuman$^*$ && 20.855 & 0.8636 & \textbf{0.0764} & - \\
        LHM && 19.275 & 0.8913 & 0.1062 & 139.566 \\
        IDOL && 19.348 & 0.9321 & 0.0919 & 201.377 \\
        Champ with GaussianAvatar && 18.264 & 0.8842 & 0.2639 & 129.413\\
        \textbf{\textit{HumanDreamer-X}} with GaussianAvatar && 19.328 & 0.8945 & 0.2578 & 132.200 \\
        Champ with Animatable gaussians && 18.908 & 0.9328 & 0.1278 & 176.836 \\
        \textbf{\textit{HumanDreamer-X}} with Animatable gaussians && \textbf{21.091} & \textbf{0.9403} & 0.0968 & \textbf{78.174}\\
        \bottomrule
    \end{tabular}
    \label{tab:recon_custom}
    \vspace{-0.3cm}
\end{table*}

\begin{figure*}[htbp]
    \centering
    \resizebox{\textwidth}{!}{
        \includegraphics{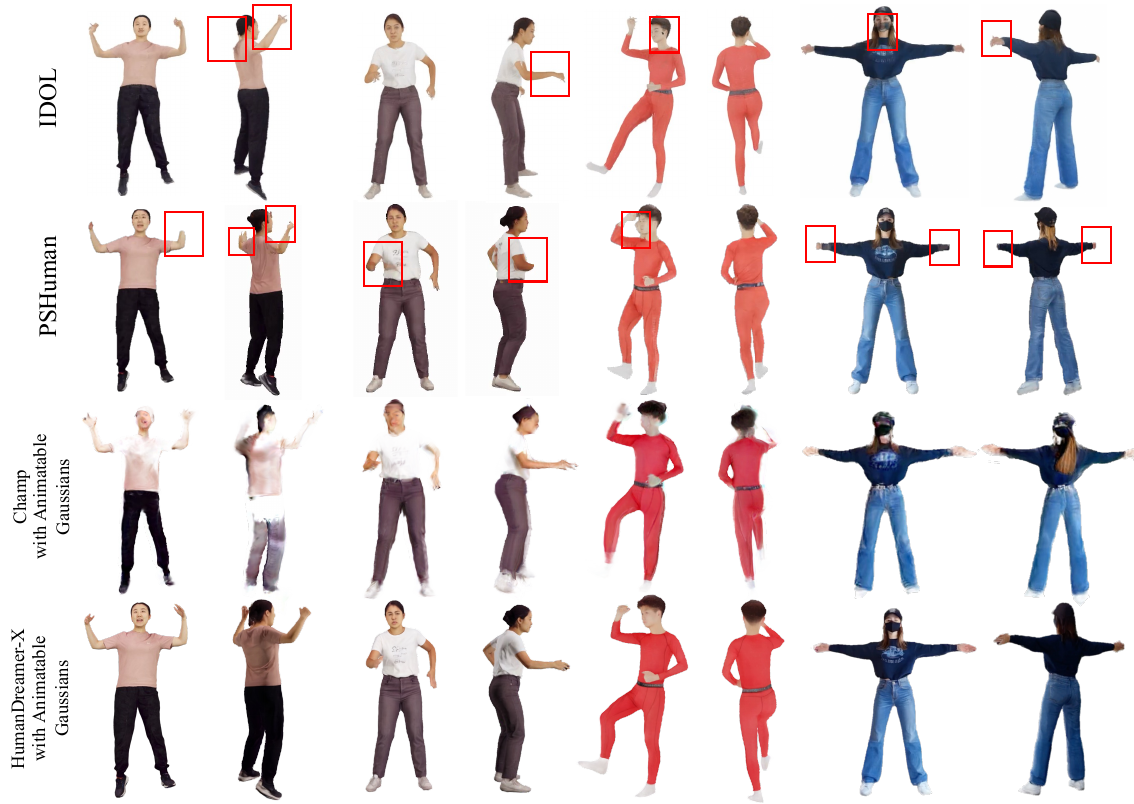}
    }
    \caption{Comparison of 3D reconstruction with SOTA methods. Best viewed with zoom-in.}
    \label{fig:recon_comp}
\end{figure*}

\noindent\textbf{Quantity comparison results on 3D reconstruction.}
To evaluate the 3D reconstruction quality, we render the reconstructed models from 18 different views and calculate the PSNR, LPIPS, SSIM, and FID metrics. We compare our approach against various baselines to demonstrate the broad applicability of our proposed framework. As shown in Tab.~\ref{tab:recon_custom}, our framework supports the integration of more advanced 3DGS backbones for enhanced reconstruction performance, thereby achieving superior results. We also conducted a visualization comparison experiment, as shown in Fig.~\ref{fig:recon_comp}. Compared to other methods, ours delivers higher detail fidelity while mitigating issues like poor identity consistency, missing hands, and limb sticking artifacts. Specifically, compared to the second-best method, PSHuman, our approach shows improvements in PSNR, SSIM, LPIPS and FID by 12.65\%, 12.07\%, 58.81\% and 18.86\% on CustomHumans.

\begin{table}[htbp]
    \centering
    \caption{Ablation Study on the Use of Coarse 3DGS.}
    \resizebox{\columnwidth}{!}{
    \begin{tabular}{lccccccc}
        \toprule
        Method  & PSNR $\uparrow$ & SSIM $\uparrow$ &LPIPS $\downarrow$ & FID $\downarrow$\\
        \midrule
        w/o coarse 3DGS & 16.824 & 0.621 & 0.3122 & 170.659\\
        w/ coarse 3DGS & \textbf{25.618} & \textbf{0.882} & \textbf{0.0687} & \textbf{87.149} \\
        \bottomrule
    \end{tabular}
    }
    \label{tab:ablation_coarse3dgs}
\end{table}

\begin{table}[htbp]
    \centering
    \caption{Ablation Study on Attention Modulation in the Generation Stage on the CustomHumans Subset.}
    \resizebox{\columnwidth}{!}{
    \begin{tabular}{lccccccc}
        \toprule
        Method  & PSNR $\uparrow$ & SSIM $\uparrow$ &LPIPS $\downarrow$ & FID $\downarrow$\\
        \midrule
        w/o attention mask (non-cyclic) & 25.514 & \textbf{0.885} & 0.0704 & 95.065\\
        w/o attention mask (cyclic) & \textbf{25.667} & 0.800 & 0.1453 & 106.705 \\
        w attention mask (cyclic)   & 25.618 & 0.882 & \textbf{0.0687} & \textbf{87.149} \\
        \bottomrule
    \end{tabular}
    }
    \label{tab:ablation_gen}
\end{table}

\begin{table}[htbp]
    \centering
    \caption{Ablation Study on Attention Modulation in the Reconstruction Stage using Animatable Gaussians~\cite{animatablegaussians}.}
    \resizebox{\columnwidth}{!}{
    \begin{tabular}{lccccccc}
        \toprule
        Method  & PSNR $\uparrow$ & SSIM $\uparrow$ &LPIPS $\downarrow$ & FID $\downarrow$\\
        \midrule
        w/o attention mask (non-cyclic) & 21.309 & 0.9398 & 0.0867 & 112.812\\
        w/o attention mask (cyclic) & 20.867 & 0.9351 & 0.0955 & 139.693 \\
        w attention mask (cyclic)   & \textbf{22.631} & \textbf{0.9458} & \textbf{0.0729} & \textbf{71.250}\\
        \bottomrule
    \end{tabular}
    }
    \label{tab:ablation_recon}
\end{table}

\subsection{Ablation Study}
We conduct an ablation study to evaluate the importance of coarse 3DGS-based multi-view reconstruction as a conditioning signal during training. Specifically, we train the generation model without using the coarse 3DGS rendering as input condition. As shown in Tab.~\ref{tab:ablation_coarse3dgs}, the significant performance drop in all metrics, demonstrates the critical role of this geometric prior in guiding high-fidelity human video generation. These results validate that coarse 3DGS rendering provides effective structural supervision, significantly improving the quality and consistency of the generated multi-view videos.

Furthermore, we train \textit{HumanFixer} on the CustomHumans dataset~\cite{customhumans}, comparing non-cyclic videos (18 frames) and cyclic videos (19 frames) with and without attention modulation, under both multi-view video generation (see Tab.~\ref{tab:ablation_gen}) and 3D reconstruction settings (see Tab.~\ref{tab:ablation_recon}). Experimental results show that while video quality degrades when transitioning from non-cyclic to cyclic inputs, the incorporation of attention modulation notably improves performance, achieving the best overall results. This confirms the effectiveness of our proposed attention mechanism in enhancing the fidelity and temporal coherence of generated avatar sequences.

\section{Limitations and Conclusions}
\textbf{Limitations}. 
Our method follows a two-stage pipeline: it first reconstructs a coarse human avatar, followed by restoration and final refinement. As described in the paper, this approach results in relatively high computational cost. Depending on the chosen 3DGS baseline, each full human avatar modeling process takes approximately 7 to 15 minutes, with the reconstruction stage alone accounting for about 90\% of the total time. Employing faster reconstruction techniques could help alleviate this computational bottleneck.

\textbf{Conclusions}.
Single-image human reconstruction is vital for digital human modeling but remains challenging due to geometric inconsistencies and limited visual fidelity. To address these issues, we propose \textit{HumanDreamer-X}, a unified framework for multi-view human generation and reconstruction. It leverages 3DGS for robust initialization and introduces \textit{HumanFixer} to refine geometry and appearance, along with an attention modulation strategy to enhance multi-view consistency. Experimental results show that our method significantly improves reconstruction quality, generalizes well to in-the-wild data, and is compatible with various backbone models.

{
    \small
    \bibliographystyle{ieeenat_fullname}
    \bibliography{PaperForReview}

@String(CVPR= {IEEE Conf. Comput. Vis. Pattern Recog.})

@String(ICCV= {Int. Conf. Comput. Vis.})

@String(ECCV= {Eur. Conf. Comput. Vis.})

@String(TOG= {ACM Trans. Graph.})

@String(CVPR  = {CVPR})

@String(ICCV  = {ICCV})

@String(ECCV  = {ECCV})

@String(TOG   = {ACM TOG})

@article{nerf,
  title={Nerf: Representing scenes as neural radiance fields for view synthesis},
  author={Mildenhall, Ben and Srinivasan, Pratul P and Tancik, Matthew and Barron, Jonathan T and Ramamoorthi, Ravi and Ng, Ren},
  journal={Communications of the ACM},
  year={2021},
}

@Article{3dgs,
      author       = {Kerbl, Bernhard and Kopanas, Georgios and Leimk{\"u}hler, Thomas and Drettakis, George},
      title        = {3D Gaussian Splatting for Real-Time Radiance Field Rendering},
      journal      = {ACM ToG},
      year         = {2023},
}

@article{worlddreamer,
  title={Worlddreamer: Towards general world models for video generation via predicting masked tokens},
  author={Wang, Xiaofeng and Zhu, Zheng and Huang, Guan and Wang, Boyuan and Chen, Xinze and Lu, Jiwen},
  journal={arXiv preprint arXiv:2401.09985},
  year={2024}
}

@inproceedings{mipnerf,
  title={Mip-nerf: A multiscale representation for anti-aliasing neural radiance fields},
  author={Barron, Jonathan T and Mildenhall, Ben and Tancik, Matthew and Hedman, Peter and Martin-Brualla, Ricardo and Srinivasan, Pratul P},
  booktitle={ICCV},
  year={2021}
}

@inproceedings{zipnerf,
  title={Zip-nerf: Anti-aliased grid-based neural radiance fields},
  author={Barron, Jonathan T and Mildenhall, Ben and Verbin, Dor and Srinivasan, Pratul P and Hedman, Peter},
  booktitle={ICCV},
  year={2023}
}

@inproceedings{mipgs,
  title={Mip-splatting: Alias-free 3d gaussian splatting},
  author={Yu, Zehao and Chen, Anpei and Huang, Binbin and Sattler, Torsten and Geiger, Andreas},
  booktitle={CVPR},
  year={2024}
}

@article{yan2021videogpt,
  title={Videogpt: Video generation using vq-vae and transformers},
  author={Yan, Wilson and Zhang, Yunzhi and Abbeel, Pieter and Srinivas, Aravind},
  journal={arXiv preprint arXiv:2104.10157},
  year={2021}
}

@article{svd,
  title={Stable video diffusion: Scaling latent video diffusion models to large datasets},
  author={Blattmann, Andreas and Dockhorn, Tim and Kulal, Sumith and Mendelevitch, Daniel and Kilian, Maciej and Lorenz, Dominik and Levi, Yam and English, Zion and Voleti, Vikram and Letts, Adam and others},
  journal={arXiv preprint arXiv:2311.15127},
  year={2023}
}

@article{ma2024latte,
  title={Latte: Latent diffusion transformer for video generation},
  author={Ma, Xin and Wang, Yaohui and Jia, Gengyun and Chen, Xinyuan and Liu, Ziwei and Li, Yuan-Fang and Chen, Cunjian and Qiao, Yu},
  journal={arXiv preprint arXiv:2401.03048},
  year={2024}
}

@inproceedings{videoldm,
  title={Align your latents: High-resolution video synthesis with latent diffusion models},
  author={Blattmann, Andreas and Rombach, Robin and Ling, Huan and Dockhorn, Tim and Kim, Seung Wook and Fidler, Sanja and Kreis, Karsten},
  booktitle={CVPR},
  year={2023}
}

@article{yang2024cogvideox,
  title={CogVideoX: Text-to-Video Diffusion Models with An Expert Transformer},
  author={Yang, Zhuoyi and Teng, Jiayan and Zheng, Wendi and Ding, Ming and Huang, Shiyu and Xu, Jiazheng and Yang, Yuanming and Hong, Wenyi and Zhang, Xiaohan and Feng, Guanyu and others},
  journal={arXiv preprint arXiv:2408.06072},
  year={2024}
}

@article{zhu2024sora,
  title={Is sora a world simulator? a comprehensive survey on general world models and beyond},
  author={Zhu, Zheng and Wang, Xiaofeng and Zhao, Wangbo and Min, Chen and Deng, Nianchen and Dou, Min and Wang, Yuqi and Shi, Botian and Wang, Kai and Zhang, Chi and others},
  journal={arXiv preprint arXiv:2405.03520},
  year={2024}
}

@inproceedings{sd,
  title={High-resolution image synthesis with latent diffusion models},
  author={Rombach, Robin and Blattmann, Andreas and Lorenz, Dominik and Esser, Patrick and Ommer, Bj{\"o}rn},
  booktitle={CVPR},
  year={2022}
}

@inproceedings{deformablegs,
  title={Deformable 3d gaussians for high-fidelity monocular dynamic scene reconstruction},
  author={Yang, Ziyi and Gao, Xinyu and Zhou, Wen and Jiao, Shaohui and Zhang, Yuqing and Jin, Xiaogang},
  booktitle={CVPR},
  year={2024}
}

@inproceedings{clip,
  title={Learning transferable visual models from natural language supervision},
  author={Radford, Alec and Kim, Jong Wook and Hallacy, Chris and Ramesh, Aditya and Goh, Gabriel and Agarwal, Sandhini and Sastry, Girish and Askell, Amanda and Mishkin, Pamela and Clark, Jack and others},
  booktitle={ICML},
  year={2021},
}

@inproceedings{champ,
      title={Champ: Controllable and Consistent Human Image Animation with 3D Parametric Guidance},
      author={Shenhao Zhu and Junming Leo Chen and Zuozhuo Dai and Yinghui Xu and Xun Cao and Yao Yao and Hao Zhu and Siyu Zhu},
      booktitle={ECCV},
      year={2024}
}

@inproceedings{animatablegaussians,
  title={Animatable Gaussians: Learning Pose-dependent Gaussian Maps for High-fidelity Human Avatar Modeling},
  author={Li, Zhe and Zheng, Zerong and Wang, Lizhen and Liu, Yebin},
  booktitle={CVPR},
  year={2024}
}

@article{anigs,
  title={AniGS: Animatable Gaussian Avatar from a Single Image with Inconsistent Gaussian Reconstruction},
  author={Qiu, Lingteng and Zhu, Shenhao and Zuo, Qi and Gu, Xiaodong and Dong, Yuan and Zhang, Junfei and Xu, Chao and Li, Zhe and Yuan, Weihao and Bo, Liefeng and others},
  journal={arXiv preprint arXiv:2412.02684},
  year={2024}
}

@article{pshuman,
  title={PSHuman: Photorealistic Single-view Human Reconstruction using Cross-Scale Diffusion},
  author={Li, Peng and Zheng, Wangguandong and Liu, Yuan and Yu, Tao and Li, Yangguang and Qi, Xingqun and Li, Mengfei and Chi, Xiaowei and Xia, Siyu and Xue, Wei and others},
  journal={arXiv preprint arXiv:2409.10141},
  year={2024}
}

@inproceedings{animateanyone,
  title={Animate anyone: Consistent and controllable image-to-video synthesis for character animation},
  author={Hu, Li},
  booktitle={CVPR},
  year={2024}
}

@InProceedings{function4d,
title={Function4D: Real-time Human Volumetric Capture from Very Sparse Consumer RGBD Sensors},
author={Yu, Tao and Zheng, Zerong and Guo, Kaiwen and Liu, Pengpeng and Dai, Qionghai and Liu, Yebin},
booktitle={CVPR},
year={2021}
}

@inproceedings{customhumans,
    title={Learning Locally Editable Virtual Humans},
    author={Hsuan-I Ho, Lixin Xue, Jie Song and Otmar Hilliges},
    booktitle={CVPR},
    year={2023}
}

@inproceedings{en3d,
  title={En3d: An enhanced generative model for sculpting 3d humans from 2d synthetic data},
  author={Men, Yifang and Lei, Biwen and Yao, Yuan and Cui, Miaomiao and Lian, Zhouhui and Xie, Xuansong},
  booktitle={CVPR},
  year={2024}
}

@inproceedings{sith,
  title={Sith: Single-view textured human reconstruction with image-conditioned diffusion},
  author={Ho, I and Song, Jie and Hilliges, Otmar and others},
  booktitle={CVPR},
  year={2024}
}

@article{mimicmotion,
  title={Mimicmotion: High-quality human motion video generation with confidence-aware pose guidance},
  author={Zhang, Yuang and Gu, Jiaxi and Wang, Li-Wen and Wang, Han and Cheng, Junqi and Zhu, Yuefeng and Zou, Fangyuan},
  journal={arXiv preprint arXiv:2406.19680},
  year={2024}
}

@article{unianimate,
  title={UniAnimate: Taming Unified Video Diffusion Models for Consistent Human Image Animation},
  author={Wang, Xiang and Zhang, Shiwei and Gao, Changxin and Wang, Jiayu and Zhou, Xiaoqiang and Zhang, Yingya and Yan, Luxin and Sang, Nong},
  journal={arXiv preprint arXiv:2406.01188},
  year={2024}
}

@article{animatex,
  title={Animate-X: Universal Character Image Animation with Enhanced Motion Representation},
  author={Tan, Shuai and Gong, Biao and Wang, Xiang and Zhang, Shiwei and Zheng, Dandan and Zheng, Ruobing and Zheng, Kecheng and Chen, Jingdong and Yang, Ming},
  journal={arXiv preprint arXiv:2410.10306},
  year={2024}
}

@inproceedings{gaussianavatar,
        title={GaussianAvatar: Towards Realistic Human Avatar Modeling from a Single Video via Animatable 3D Gaussians},
        author={Hu, Liangxiao and Zhang, Hongwen and Zhang, Yuxiang and Zhou, Boyao and Liu, Boning and Zhang, Shengping and Nie, Liqiang},
        booktitle={CVPR},
        year={2024}
}

@inproceedings{gauhuman,
  title={Gauhuman: Articulated gaussian splatting from monocular human videos},
  author={Hu, Shoukang and Hu, Tao and Liu, Ziwei},
  booktitle={CVPR},
  year={2024}
}

@inproceedings{gart,
  title={Gart: Gaussian articulated template models},
  author={Lei, Jiahui and Wang, Yufu and Pavlakos, Georgios and Liu, Lingjie and Daniilidis, Kostas},
  booktitle={CVPR},
  year={2024}
}

@inproceedings{instantavatar,
  title={Instantavatar: Learning avatars from monocular video in 60 seconds},
  author={Jiang, Tianjian and Chen, Xu and Song, Jie and Hilliges, Otmar},
  booktitle={CVPR},
  year={2023}
}

@inproceedings{humannerf,
  title={Humannerf: Free-viewpoint rendering of moving people from monocular video},
  author={Weng, Chung-Yi and Curless, Brian and Srinivasan, Pratul P and Barron, Jonathan T and Kemelmacher-Shlizerman, Ira},
  booktitle={CVPR},
  year={2022}
}

@inproceedings{sifu,
  title={Sifu: Side-view conditioned implicit function for real-world usable clothed human reconstruction},
  author={Zhang, Zechuan and Yang, Zongxin and Yang, Yi},
  booktitle={CVPR},
  year={2024}
}

@article{magicman,
  title={Magicman: Generative novel view synthesis of humans with 3d-aware diffusion and iterative refinement},
  author={He, Xu and Li, Xiaoyu and Kang, Di and Ye, Jiangnan and Zhang, Chaopeng and Chen, Liyang and Gao, Xiangjun and Zhang, Han and Wu, Zhiyong and Zhuang, Haolin},
  journal={arXiv preprint arXiv:2408.14211},
  year={2024}
}

@article{idol,
  title={IDOL: Instant Photorealistic 3D Human Creation from a Single Image},
  author={Zhuang, Yiyu and Lv, Jiaxi and Wen, Hao and Shuai, Qing and Zeng, Ailing and Zhu, Hao and Chen, Shifeng and Yang, Yujiu and Cao, Xun and Liu, Wei},
  journal={arXiv preprint arXiv:2412.14963},
  year={2024}
}

@article{charactergen,
  title={Charactergen: Efficient 3d character generation from single images with multi-view pose canonicalization},
  author={Peng, Hao-Yang and Zhang, Jia-Peng and Guo, Meng-Hao and Cao, Yan-Pei and Hu, Shi-Min},
  journal={ACM ToG},
  year={2024},
}

@incollection{smpl,
  title={SMPL: A skinned multi-person linear model},
  author={Loper, Matthew and Mahmood, Naureen and Romero, Javier and Pons-Moll, Gerard and Black, Michael J},
  booktitle={Seminal Graphics Papers},
  year={2023}
}

@misc{gas,
      title={GAS: Generative Avatar Synthesis from a Single Image}, 
      author={Yixing Lu and Junting Dong and Youngjoong Kwon and Qin Zhao and Bo Dai and Fernando De la Torre},
      year={2025},
      eprint={2502.06957},
      archivePrefix={arXiv},
      primaryClass={cs.CV},
      url={https://arxiv.org/abs/2502.06957}, 
}

@inproceedings{magicanimate,
  title={Magicanimate: Temporally consistent human image animation using diffusion model},
  author={Xu, Zhongcong and Zhang, Jianfeng and Liew, Jun Hao and Yan, Hanshu and Liu, Jia-Wei and Zhang, Chenxu and Feng, Jiashi and Shou, Mike Zheng},
  booktitle={CVPR},
  year={2024}
}

@inproceedings{densepose,
  title={Densepose: Dense human pose estimation in the wild},
  author={G{\"u}ler, R{\i}za Alp and Neverova, Natalia and Kokkinos, Iasonas},
  booktitle={CVPR},
  year={2018}
}

@article{disco,
  title={Disco: Disentangled control for realistic human dance generation},
  author={Wang, Tan and Li, Linjie and Lin, Kevin and Zhai, Yuanhao and Lin, Chung-Ching and Yang, Zhengyuan and Zhang, Hanwang and Liu, Zicheng and Wang, Lijuan},
  journal={arXiv preprint arXiv:2307.00040},
  year={2023}
}

@misc{vae,
      title={Auto-Encoding Variational Bayes}, 
      author={Diederik P Kingma and Max Welling},
      year={2022},
      eprint={1312.6114},
      archivePrefix={arXiv},
      primaryClass={stat.ML},
      url={https://arxiv.org/abs/1312.6114}, 
}

@article{mipnerf360,
    title={Mip-NeRF 360: Unbounded Anti-Aliased Neural Radiance Fields},
    author={Jonathan T. Barron and Ben Mildenhall and 
            Dor Verbin and Pratul P. Srinivasan and Peter Hedman},
    journal={CVPR},
    year={2022}
}

@InProceedings{Hu_2021_CVPR,
    author    = {Hu, Tao and Wang, Liwei and Xu, Xiaogang and Liu, Shu and Jia, Jiaya},
    title     = {Self-Supervised 3D Mesh Reconstruction From Single Images},
    booktitle = {CVPR},
    year      = {2021},
}

@article{meshanything,
  title={Meshanything: Artist-created mesh generation with autoregressive transformers},
  author={Chen, Yiwen and He, Tong and Huang, Di and Ye, Weicai and Chen, Sijin and Tang, Jiaxiang and Chen, Xin and Cai, Zhongang and Yang, Lei and Yu, Gang and others},
  journal={arXiv preprint arXiv:2406.10163},
  year={2024}
}

@inproceedings{lu2024scaffold,
  title={Scaffold-gs: Structured 3d gaussians for view-adaptive rendering},
  author={Lu, Tao and Yu, Mulin and Xu, Linning and Xiangli, Yuanbo and Wang, Limin and Lin, Dahua and Dai, Bo},
  booktitle={CVPR},
  year={2024}
}

@article{mamba,
  title={Mamba: Linear-time sequence modeling with selective state spaces},
  author={Gu, Albert and Dao, Tri},
  journal={arXiv preprint arXiv:2312.00752},
  year={2023}
}

@article{mimo,
  title={Mimo: Controllable character video synthesis with spatial decomposed modeling},
  author={Men, Yifang and Yao, Yuan and Cui, Miaomiao and Bo, Liefeng},
  journal={arXiv preprint arXiv:2409.16160},
  year={2024}
}

@article{animateanyone2,
      title={Animate Anyone 2: High-Fidelity Character Image Animation with Environment Affordance},
      author={Li Hu and Guangyuan Wang and Zhen Shen and Xin Gao and Dechao Meng and Lian Zhuo and Peng Zhang and Bang Zhang and Liefeng Bo},
      journal={arXiv preprint arXiv:2502.06145},
      year={2025}
}

@article{eva3d,
    title={EVA3D: Compositional 3D Human Generation from 2D Image Collections},
    author={Hong, Fangzhou and Chen, Zhaoxi and Lan, Yushi and Pan, Liang and Liu, Ziwei},
    journal={arXiv preprint arXiv:2210.04888},
    year={2022}
}

@inproceedings{3dgsavatar,
  title={3dgs-avatar: Animatable avatars via deformable 3d gaussian splatting},
  author={Qian, Zhiyin and Wang, Shaofei and Mihajlovic, Marko and Geiger, Andreas and Tang, Siyu},
  booktitle={CVPR},
  year={2024}
}

@article{avatarrex,
  title={Avatarrex: Real-time expressive full-body avatars},
  author={Zheng, Zerong and Zhao, Xiaochen and Zhang, Hongwen and Liu, Boning and Liu, Yebin},
  journal={ACM ToG},
  year={2023},
}

@article{evahuman,
  title={Expressive gaussian human avatars from monocular rgb video},
  author={Hu, Hezhen and Fan, Zhiwen and Wu, Tianhao and Xi, Yihan and Lee, Seoyoung and Pavlakos, Georgios and Wang, Zhangyang and others},
  journal={NeurIPS},
  year={2025}
}

@inproceedings{hugs,
  title={Human gaussian splatting: Real-time rendering of animatable avatars},
  author={Moreau, Arthur and Song, Jifei and Dhamo, Helisa and Shaw, Richard and Zhou, Yiren and P{\'e}rez-Pellitero, Eduardo},
  booktitle={CVPR},
  year={2024}
}

@article{humanrf,
  title={Humanrf: High-fidelity neural radiance fields for humans in motion},
  author={I{\c{s}}{\i}k, Mustafa and R{\"u}nz, Martin and Georgopoulos, Markos and Khakhulin, Taras and Starck, Jonathan and Agapito, Lourdes and Nie{\ss}ner, Matthias},
  journal={ACM ToG},
  year={2023},
}

@article{humansplat,
  title={Humansplat: Generalizable single-image human gaussian splatting with structure priors},
  author={Pan, Panwang and Su, Zhuo and Lin, Chenguo and Fan, Zhen and Zhang, Yongjie and Li, Zeming and Shen, Tingting and Mu, Yadong and Liu, Yebin},
  journal={NeurIPS},
  year={2024}
}

@inproceedings{ag3d,
  title={Ag3d: Learning to generate 3d avatars from 2d image collections},
  author={Dong, Zijian and Chen, Xu and Yang, Jinlong and Black, Michael J and Hilliges, Otmar and Geiger, Andreas},
  booktitle={ICCV},
  year={2023}
}

@inproceedings{eg3d,
  title={Efficient geometry-aware 3d generative adversarial networks},
  author={Chan, Eric R and Lin, Connor Z and Chan, Matthew A and Nagano, Koki and Pan, Boxiao and De Mello, Shalini and Gallo, Orazio and Guibas, Leonidas J and Tremblay, Jonathan and Khamis, Sameh and others},
  booktitle={CVPR},
  year={2022}
}

@inproceedings{emo,
  title={Emo: Emote portrait alive generating expressive portrait videos with audio2video diffusion model under weak conditions},
  author={Tian, Linrui and Wang, Qi and Zhang, Bang and Bo, Liefeng},
  booktitle={ECCV},
  pages={244--260},
  year={2024},
  organization={Springer}
}

@inproceedings{followyouremoji,
  title={Follow-your-emoji: Fine-controllable and expressive freestyle portrait animation},
  author={Ma, Yue and Liu, Hongyu and Wang, Hongfa and Pan, Heng and He, Yingqing and Yuan, Junkun and Zeng, Ailing and Cai, Chengfei and Shum, Heung-Yeung and Liu, Wei and others},
  booktitle={SIGGRAPH Asia},
  year={2024}
}

@article{aniportrait,
  title={Aniportrait: Audio-driven synthesis of photorealistic portrait animation},
  author={Wei, Huawei and Yang, Zejun and Wang, Zhisheng},
  journal={arXiv preprint arXiv:2403.17694},
  year={2024}
}

@inproceedings{controlnet,
  title={Adding conditional control to text-to-image diffusion models},
  author={Zhang, Lvmin and Rao, Anyi and Agrawala, Maneesh},
  booktitle={ICCV},
  year={2023}}

@article{ipadapter,
  title={Ip-adapter: Text compatible image prompt adapter for text-to-image diffusion models},
  author={Ye, Hu and Zhang, Jun and Liu, Sibo and Han, Xiao and Yang, Wei},
  journal={arXiv preprint arXiv:2308.06721},
  year={2023}
}

@inproceedings{arcface,
title={ArcFace: Additive Angular Margin Loss for Deep Face Recognition},
author={Deng, Jiankang and Guo, Jia and Niannan, Xue and Zafeiriou, Stefanos},
booktitle={CVPR},
year={2019}
}

@article{lhm,
title={LHM: Large Animatable Human Reconstruction Model from a Single Image in Seconds},
author={Lingteng Qiu and Xiaodong Gu and Peihao Li  and Qi Zuo
  and Weichao Shen and Junfei Zhang and Kejie Qiu and Weihao Yuan
  and Guanying Chen and Zilong Dong and Liefeng Bo 
  },
booktitle={arXiv preprint arXiv:2503.10625},
year={2025}
}

@ARTICLE{pamir,
author={Zheng, Zerong and Yu, Tao and Liu, Yebin and Dai, Qionghai},
journal={IEEE Transactions on Pattern Analysis and Machine Intelligence},
title={PaMIR: Parametric Model-Conditioned Implicit Representation for Image-based Human Reconstruction},
year={2021},
volume={},
number={},
pages={1-1},
doi={10.1109/TPAMI.2021.3050505}
}

@inproceedings{pifu,
  title={Pifu: Pixel-aligned implicit function for high-resolution clothed human digitization},
  author={Saito, Shunsuke and Huang, Zeng and Natsume, Ryota and Morishima, Shigeo and Kanazawa, Angjoo and Li, Hao},
  booktitle={Proceedings of the IEEE/CVF international conference on computer vision},
  pages={2304--2314},
  year={2019}
}

@ARTICLE{SparseFusion,
  author={Zuo, Xinxin and Wang, Sen and Zheng, Jiangbin and Yu, Weiwei and Gong, Minglun and Yang, Ruigang and Cheng, Li},
  journal={IEEE Transactions on Multimedia}, 
  title={SparseFusion: Dynamic Human Avatar Modeling From Sparse RGBD Images}, 
  year={2021},
  volume={23},
  number={},
  pages={1617-1629},
  doi={10.1109/TMM.2020.3001506}}

@ARTICLE{9787789,
  author={Sun, Qingping and Xiao, Yi and Zhang, Jie and Zhou, Shizhe and Leung, Chi-Sing and Su, Xin},
  journal={IEEE Transactions on Multimedia}, 
  title={A Local Correspondence-Aware Hybrid CNN-GCN Model for Single-Image Human Body Reconstruction}, 
  year={2023},
  volume={25},
  number={},
  pages={4679-4690},
  doi={10.1109/TMM.2022.3180218}}

@ARTICLE{Text2Avatar,
  author={Kwon, Yong-Hoon and Yoon, Ju Hong and Park, Min-Gyu},
  journal={IEEE Transactions on Multimedia}, 
  title={Text2Avatar: Articulated 3D Avatar Creation With Text Instructions}, 
  year={2025},
  volume={27},
  number={},
  pages={3797-3806},
  doi={10.1109/TMM.2025.3535293}}

@ARTICLE{10229247,
  author={Shen, Shuai and Li, Wanhua and Huang, Xiaoke and Zhu, Zheng and Zhou, Jie and Lu, Jiwen},
  journal={IEEE Transactions on Multimedia}, 
  title={SD-NeRF: Towards Lifelike Talking Head Animation via Spatially-Adaptive Dual-Driven NeRFs}, 
  year={2024},
  volume={26},
  number={},
  pages={3221-3234},
  doi={10.1109/TMM.2023.3308441}}

@ARTICLE{10543061,
  author={Zhang, Shufang and Ni, Minxue and Chen, Shuai and Wang, Lei and Ding, Wenxin and Liu, Yuhong},
  journal={IEEE Transactions on Multimedia}, 
  title={A Two-Stage Personalized Virtual Try-On Framework With Shape Control and Texture Guidance}, 
  year={2024},
  volume={26},
  number={},
  pages={10225-10236},
  doi={10.1109/TMM.2024.3405718}}
}

\end{document}